\documentclass[12pt]{article}

\usepackage{adjustbox}
\usepackage{amsmath}
\usepackage{amssymb}
\usepackage{dutchcal}
\usepackage{enumitem}
\usepackage{float}
\usepackage[T1]{fontenc}
\usepackage{graphicx}
\usepackage{hhline}
\usepackage[utf8]{inputenc}
\usepackage{mathtools}
\usepackage{multicol}
\usepackage{multirow}
\usepackage{wasysym}
\usepackage[paperheight=29.7cm,paperwidth=20.99cm,left=3.0cm,right=3.0cm,top=2.5cm,bottom=2.5cm]{geometry}
\usepackage{xurl}

\begin{document}

\begin{center}
\textbf{A Fully Automated Pipeline Using Swin Transformers for Deep Learning-Based Blood Segmentation on Head CT Scans After Aneurysmal Subarachnoid Hemorrhage}
\end{center}

\vspace{1\baselineskip}
\begin{center}
\textit{Sergio García García, MD., Ph.D.,\textsuperscript{1,} † Santiago Cepeda, MD., Ph.D.,\textsuperscript{1,} † Ignacio Arrese MD., Ph.D.,\textsuperscript{1  }Rosario Sarabia, MD., Ph.D.\textsuperscript{1}}
\end{center}

\vspace{1\baselineskip}
\begin{enumerate}
	\item Department of Neurosurgery, Río Hortega University Hospital, Valladolid, Spain.

\end{enumerate}
\begin{center}
†\textit{ These authors contributed equally to the research presented herein.}
\end{center}

\vspace{2\baselineskip}
\textbf{ABSTRACT}

\vspace{1\baselineskip}
\textbf{Background: }Accurate volumetric assessment of spontaneous subarachnoid hemorrhage (SAH) is a labor-intensive task performed with current manual and semiautomatic methods that might be relevant for its clinical and prognostic implications.\textbf{ }In the present research, we sought to develop and validate an artificial intelligence-driven, fully automated blood segmentation tool for SAH patients via noncontrast computed tomography (NCCT) scans employing a transformer-based Swin UNETR architecture.

\textbf{Methods: }We retrospectively analyzed NCCT scans from patients with confirmed aneurysmal subarachnoid hemorrhage (aSAH) utilizing the Swin UNETR for segmentation. The performance of the proposed method was evaluated against manually segmented ground truth data using metrics such as Dice score, intersection over union (IoU), the volumetric similarity index (VSI), the symmetric average surface distance (SASD), and sensitivity and specificity. A validation cohort from an external institution was included to test the generalizability of the model.

\textbf{Results: }The model demonstrated high accuracy with robust performance metrics across the internal and external validation cohorts. Notably, it achieved high Dice coefficient (0.873 $\pm$ 0.097), IoU (0.810 $\pm$ 0.092), VSI (0.840 $\pm$ 0.131), sensitivity (0.821 $\pm$ 0.217) and specificity (0.996 $\pm$ 0.004) values and a low SASD (1.866 $\pm$ 2.910), suggesting proficiency in segmenting blood in SAH patients. The model's efficiency was reflected in its processing speed, indicating potential for real-time applications.

\textbf{Conclusions: }Our Swin UNETR-based model offers significant advances in the automated segmentation of blood after aSAH on NCCT images. Despite the computational intensity, the model operates effectively on standard hardware with a user-friendly interface, facilitating broader clinical adoption. Further validation across diverse datasets is warranted to confirm its clinical reliability.

\vspace{6\baselineskip}
\textbf{INTRODUCTION}

Spontaneous subarachnoid hemorrhage (SAH) is a devastating disease frequently caused by the rupture of an intracranial aneurysm (1). Aneurysmal SAH is associated with high mortality and morbidity rates despite optimal management . Various prognostic factors have been identified either as previous conditions determining a more threatening course and poorer outcome, such as age or hypertension, or as developing complications that might worsen the final clinical result, such as seizures, vasospasm, delayed cerebral ischemia and hydrocephalus (2-8). Some of these factors might be identified on admission, while others may appear days after the onset of SAH. Clinical research initiatives have addressed their efforts to identify early factors that could predict which patients are at higher risk of a poor outcome to determine who might benefit from a more aggressive or compassionate management strategy.

The amount of bleeding has been shown to be significantly associated with mortality, poor outcomes and other complications, such as delayed cerebral ischemia or hydrocephalus(5, 7, 9-11). Calculating the volume of blood in the SAH cohort might be cumbersome and time consuming. Various methods and software programs have been implemented for blood segmentation from noncontrast computed tomography (NCCT) scans. Manual, automatic and semiautomatic methods are available with different workflows, computational requirements and accuracies(9, 11-13).

Deep learning (DL), specifically convolutional neural networks (CNNs), has been recently used to fully automate tedious tasks such as anatomic segmentation. A U-shaped CNN, for instance, U-Net, whose architecture consists of an encoder-decoder structure, has provided excellent results in medical semantic segmentation. However, their performance decreases in the segmentation of structures with intricate shapes due to the locality of the receptive field of convolutional layers. Therefore, CNNs might not be the ideal method for successfully segmenting hemorrhages in SAH patients due to their diffuse, sometimes patchy and highly variable distribution. Transformer-based models have overcome some of these limitations because of their self-attention mechanisms, which allow them to consider the entire context of the image, enabling them to record long-range dependencies in the data(14). Swin UNETR is a hybrid model that synergizes the architectural strengths of Swin transformers with the segmentation ability of U-Net structures. As a U-shaped CNN, the Swin U-TIRE consists of encoder and decoder paths composed of multiple layers of self-attention and feedforward neural networks.

In this investigation, we sought to create, implement and test a fully automated method to obtain volumetric segmentation of the hemorrhage present on the NCCT scans of patients with aneurismal SAH using a transformer-based approach.

\vspace{1\baselineskip}
\textbf{METHODOLOGY}

The present article adheres to the Strengthening the Reporting of Observational Studies in Epidemiology (STROBE) guidelines(15) and the CheckList for Artificial Intelligence in Medical Imaging (CLAIM) guidelines(16). The research protocol was approved by the Institutional Review Board (22-PI180). Considering the retrospective, noninterventional nature of the study, the IRB waived the requirement to obtain informed consent.

\vspace{1\baselineskip}
\textbf{\textit{Dataset}}

The dataset is composed of a consecutive series of patients admitted to our hospital with a confirmed diagnosis of aneurysmal SAH between 2016 and 2022. SAH diagnosis was based on compatible clinical signs and a positive NCCT scan without a previous history of a recent traumatic event. The etiology of the aneurysm was confirmed by either angio-CT or digital subtraction angiography. Patients with SAH from other causes were excluded. An external cohort of 10 patients with a confirmed diagnosis of aneurysmal SAH from a referring institution was also included to test the robustness of the model for patients from another institution with different image acquisition protocols.

\vspace{1\baselineskip}
\textbf{\textit{Variables}}

Demographic, clinical and radiological data were collected for classification purposes. Blood volume in cm\textsuperscript{3} was calculated.

\vspace{1\baselineskip}
\textbf{\textit{Image Acquisition}}

CT scans were acquired on a Siemens Somaton CT scanner (Munich, Germany) (Supplementary Table 1). The external test cohort scans were acquired on a Siemens Somatom or on Philips Ingenuity CT Scanners (Koninklijke, The Netherlands).

\vspace{1\baselineskip}
\textbf{\textit{Image Preprocessing Overview}}

NCCT files, sourced from the PACS in DICOM format, underwent several transformation and normalization procedures. The initial phase involved converting these images to the NIfTI format using the dicom2niix tool, version v1.0.20220720, available at https://github.com/rordenlab/dcm2niix/releases/tag/v1.0.20220720 (accessed on March 20\textsuperscript{th,} 2023). Notably, due to gantry tilt adjustments in some CT scans, the corrected output from the dcm2niix was selected for subsequent processing to ensure accurate 3D operation.

Traditional CT analyses employ Hounsfield units (HUs); however, interfacing with MRI pipelines can pose challenges due to negative HU values. To circumvent this, images were transitioned to Cormack units, starting at zero, using the Clinical Toolbox for SPM found at \url{https://github.com/neurolabusc/Clinical} (accessed on March 20\textsuperscript{th,} 2023).

Further preprocessing involved brain extraction via the Brain Extraction Tool (BET) from FSL v6.0, accessible at \url{https://fsl.fmrib.ox.ac.uk/fsl/fslwiki/BET/UserGuide} (accessed on March \textsuperscript{22,} 2023). The final step involved registering images to a CT template with dimensions of 1 mm $\times$ 1 mm $\times$ 1 mm and a size of [193, 229, 193]. This registration utilized diffeomorphic registrations, specifically symmetric normalization (SyN), from ANTs available at \url{https://github.com/ANTsX/ANTs} (accessed on March \textsuperscript{23,} 2023).

\vspace{1\baselineskip}
\textbf{\textit{Ground Truth Input. Manual segmentation}}

The resulting NIFTI files of 100 patients were manually segmented using a semiautomatic approach based on an edge-based snake evolution method (ITK-Snap version 4.0 (17)). All segmentation procedures were independently performed by two neurosurgeons (S.G.G. and S.C.) with more than 10 years of experience in CT imaging applied for hemorrhagic strokes.  Segmentations were reviewed and corrected by agreement in cases of mismatch. The training, validation and test datasets were randomly split into 70-10-20 distributions. Additionally, the NCCT scans of 10 patients from another institution were segmented following the same procedure.

\vspace{1\baselineskip}
\textbf{\textit{Architectural overview}}

Swin UNETR represents a cutting-edge approach to semantic segmentation in medical imaging. Traditional techniques, primarily based on fully convolutional neural networks such as UNET, have been prevalent due to their ability to handle image-based tasks. However, fully convolutional neural networks (FCNs) occasionally face challenges with extensive long-range dependencies because of their inherent convolutional nature. In contrast, the Swin UNETR leverages the strengths of transformer architectures, as it is renowned for its proficiency in managing long-range information in fields beyond imaging. By reconceptualizing the segmentation task as a sequence prediction, Swin UNETR transforms diverse input data into a streamlined sequence. These data are then channeled through a multiresolution Swin transformer encoder, which, when paired with its corresponding CNN decoder, ensures a comprehensive and precise segmentation outcome. The architecture is delineated as follows:

\vspace{1\baselineskip}
\begin{enumerate}[label*=\arabic*.]
	\item \textbf{Embedding Sequence Formation}: The architecture initiates with the conversion of diverse input modalities into a linear sequence of embeddings, streamlining the subsequent processing stages.

	\item \textbf{Hierarchical Transformer Encoding}: Central to the Swin UNETR is its hierarchical Swin transformer encoder. This encoder adept at multiresolution feature extraction. Notably, self-attention mechanisms employ shifted windowing strategies, optimizing the capture of spatial dependencies across varying scales.

	\item \textbf{Decoder Integration}: Each resolution level of the Swin transformer encoder is interfaced with a CNN-based decoder. This integration is fortified by skip connections, ensuring a bidirectional flow of spatial information, thus preserving the granularity of details across the encoding and decoding phases.

\end{enumerate}
In summary, the Swin UNETR offers a sophisticated framework adept at handling intricate segmentation tasks by seamlessly intertwining transformer-based feature extraction with traditional convolutional processing.

We employed the soft Dice loss function for our methodology, which was computed on a voxelwise basis and formulated as follows:

\begin{equation}
\mathcal{L}(G, Y) = 1 - \frac{2}{J} \sum_{j=1}^{J} \left( \frac{\sum_{i=1}^{I} G_{i,j} Y_{i,j}}{\sum_{i=1}^{I} G_{i,j}^2 + \sum_{i=1}^{I} Y_{i,j}^2} \right)
\end{equation}

Here, \textit{I} signifies the total number of voxels, \textit{J} represents the number of classes, \textit{Y\textsubscript{i,j}} and \textit{G\textsubscript{i,j}} correspond to the probability of output and the one-hot encoded ground truth for class \textit{j} at voxel \textit{i}, respectively. This loss function additionally captures the voxel-level discrepancies between the predicted and ground truth class distributions.

\vspace{1\baselineskip}
\textbf{\textit{Evaluation metrics}}

To assess the accuracy of our segmentation predictions, we employed both the Dice score and the intersection over union (IoU) overlap metric. The following notations were used throughout the evaluations:

\begin{itemize}
    \item $G_i$: Represents the ground truth value for the \textit{i\textsuperscript{th}} voxel.
    \item $P_i$: Denotes the predicted value for the \textit{i\textsuperscript{th}} voxel.
    \item $G_0$: Refers to the set of surface points from the ground truth.
    \item $P_0$: Stands for the set of surface points from our predictions.
\end{itemize}

The Dice coefficient, used as a performance metric in this context, is defined as:
\begin{equation}
\text{Dice}(G, P) = \frac{2 \times \sum_{i} G_{i} P_{i}}{\sum_{i} G_{i} + \sum_{i} P_{i}}
\end{equation}

\vspace{1\baselineskip}
\begin{equation}
\text{IoU}(G, P) = \frac{\sum_{i} G_{i} P_{i}}{\sum_{i} G_{i} + \sum_{i} P_{i} - \sum_{i} G_{i} P_{i}}
\end{equation}

In addition to the overlap metrics, we employed the volumetric similarity index (VSI), a quantification method that directly addresses the likeness between segments based on their volumes. This approach involves calculating the absolute volume difference between the compared segments and normalizing it by the sum of their volumes. This comparison is established by contrasting the absolute volume of the ground truth label (\textit{GTv})m with the corresponding volume within the predicted label (\textit{Pv}).

\begin{equation}
VSI = 1 - \frac{|GT_v - Pv|}{GT_v + Pv}
\end{equation}

\vspace{1\baselineskip}
By incorporating a distance metric into our analysis, we leverage the symmetric average surface distance (SASD) metric to assess the geometric concordance between a ground truth object (\textit{Gi}) and a predicted object (\textit{Pi}), accounting for bidirectional measurements. This metric quantifies the average distances between corresponding points on the surfaces of both objects, offering a comprehensive evaluation of shape dissimilarity:

\begin{equation}
SASD = \frac{1}{N} \sum_{i=1}^{N} \left( d(P_i, G) + d(G_i, P) \right)
\end{equation}

\vspace{1\baselineskip}
\textbf{\textit{Implementation details}}

Harnessing the potential of the PyTorch and MONAI frameworks, we implemented the Swin UNETR model on an NVIDIA GeForce 3090 GPU (18). A pivotal preprocessing measure involved standardizing the images to a zero mean and unit variance, concentrating particularly on nonzero voxels. Our data augmentation strategy included random flipping, zooming, precise region-of-interest cropping, three-dimensional axis mirroring, and channel-specific intensity adjustments. To optimize the computational processes, a constant GPU batch size of 1 was used. The model's configuration was defined with image dimensions set to [128, 128, 64] and pixel dimensions set to [1.5, 1.5, 2]. A singular input channel was employed due to the use of a single modality, the noncontrast head CT scan. In contrast, two output channels were designated, one corresponding to the segmented blood and the other to the background. Other model intricacies included a feature depth of 48. Probing the architecture further, layering was established in [2,2,2,2], with attention heads distributed across [3,6,12,24]. The model was subjected to comprehensive training over 300 epochs. For inference, we opted for a sliding window approach, maintaining a 0.25 voxel overlap and specifying a batch size of 4 for window slices.

\vspace{1\baselineskip}
\textbf{RESULTS}

The sample was composed of 110 patients with aneurysmal SAH. The characteristics of the sample are presented in Table 1.

\begin{table}[H]
\begin{adjustbox}{max width=\textwidth}
\begin{tabular}{p{4.35cm}p{2.96cm}p{2.2cm}p{2.52cm}p{2.93cm}}
\hline
\multicolumn{5}{|p{14.959999999999999cm}|}{\textbf{Table 1. Clinical and demographic features of the cohort of patients.}} \\ 
\hline
\multicolumn{2}{|p{7.31cm}}{\textbf{Variable}} & 
\multicolumn{1}{|p{2.96cm}}{\centering
\textbf{Training}} & 
\multicolumn{1}{|p{2.2cm}}{\centering
\textbf{Test}} & 
\multicolumn{1}{|p{2.52cm}|}{\centering
\textbf{External Test}} \\ 
\hline
\multicolumn{2}{|p{7.31cm}}{Patients} & 
\multicolumn{1}{|p{2.96cm}}{\centering
80} & 
\multicolumn{1}{|p{2.2cm}}{\centering
20} & 
\multicolumn{1}{|p{2.52cm}|}{\centering
10} \\ 
\hline
\multicolumn{2}{|p{7.31cm}}{Male} & 
\multicolumn{1}{|p{2.96cm}}{\centering
52$\%$} & 
\multicolumn{1}{|p{2.2cm}}{\centering
50$\%$} & 
\multicolumn{1}{|p{2.52cm}|}{\centering
40$\%$} \\ 
\hline
\multicolumn{2}{|p{7.31cm}}{Age (Years)} & 
\multicolumn{1}{|p{2.96cm}}{\centering
56  12} & 
\multicolumn{1}{|p{2.2cm}}{\centering
54  12} & 
\multicolumn{1}{|p{2.52cm}|}{\centering
59  14} \\ 
\hline
\multicolumn{2}{|p{7.31cm}}{HT} & 
\multicolumn{1}{|p{2.96cm}}{\centering
49$\%$} & 
\multicolumn{1}{|p{2.2cm}}{\centering
55$\%$} & 
\multicolumn{1}{|p{2.52cm}|}{\centering
30$\%$} \\ 
\hline
\multicolumn{2}{|p{7.31cm}}{Tobacco} & 
\multicolumn{1}{|p{2.96cm}}{\centering
51$\%$} & 
\multicolumn{1}{|p{2.2cm}}{\centering
45$\%$} & 
\multicolumn{1}{|p{2.52cm}|}{\centering
20$\%$} \\ 
\hline
\multicolumn{2}{|p{7.31cm}}{Aneurysms} & 
\multicolumn{1}{|p{2.96cm}}{\centering
113} & 
\multicolumn{1}{|p{2.2cm}}{\centering
18} & 
\multicolumn{1}{|p{2.52cm}|}{\centering
12} \\ 
\hline
\multicolumn{1}{|p{4.35cm}}{\multirow{2}{*}{\parbox{4.35cm}{\centering
WFNS}}} & 
\multicolumn{1}{|p{2.96cm}}{\centering
Mode} & 
\multicolumn{1}{|p{2.2cm}}{\centering
2} & 
\multicolumn{1}{|p{2.52cm}}{\centering
1} & 
\multicolumn{1}{|p{2.93cm}|}{\centering
4} \\ 
\hhline{~----}
\multicolumn{1}{|p{4.35cm}}{} & 
\multicolumn{1}{|p{2.96cm}}{\centering
Median} & 
\multicolumn{1}{|p{2.2cm}}{\centering
2} & 
\multicolumn{1}{|p{2.52cm}}{\centering
2} & 
\multicolumn{1}{|p{2.93cm}|}{\centering
4} \\ 
\hline
\multicolumn{1}{|p{4.35cm}}{\multirow{2}{*}{\parbox{4.35cm}{\centering
mFisher}}} & 
\multicolumn{1}{|p{2.96cm}}{\centering
Mode} & 
\multicolumn{1}{|p{2.2cm}}{\centering
4} & 
\multicolumn{1}{|p{2.52cm}}{\centering
4} & 
\multicolumn{1}{|p{2.93cm}|}{\centering
4} \\ 
\hhline{~----}
\multicolumn{1}{|p{4.35cm}}{} & 
\multicolumn{1}{|p{2.96cm}}{\centering
Median} & 
\multicolumn{1}{|p{2.2cm}}{\centering
4} & 
\multicolumn{1}{|p{2.52cm}}{\centering
4} & 
\multicolumn{1}{|p{2.93cm}|}{\centering
4} \\ 
\hline
\multicolumn{1}{|p{4.35cm}}{\multirow{3}{*}{\parbox{4.35cm}{\centering
Blood Volume (cm\textsuperscript{3})}}} & 
\multicolumn{1}{|p{2.96cm}}{\centering
Mean} & 
\multicolumn{1}{|p{2.2cm}}{\centering
44.67} & 
\multicolumn{1}{|p{2.52cm}}{\centering
65.95} & 
\multicolumn{1}{|p{2.93cm}|}{\centering
46.4} \\ 
\hhline{~----}
\multicolumn{1}{|p{4.35cm}}{} & 
\multicolumn{1}{|p{2.96cm}}{\centering
SD} & 
\multicolumn{1}{|p{2.2cm}}{\centering
27.5} & 
\multicolumn{1}{|p{2.52cm}}{\centering
63} & 
\multicolumn{1}{|p{2.93cm}|}{\centering
63} \\ 
\hhline{~----}
\multicolumn{1}{|p{4.35cm}}{} & 
\multicolumn{1}{|p{2.96cm}}{\centering
95$\%$CI} & 
\multicolumn{1}{|p{2.2cm}}{\centering
35.1-54.3} & 
\multicolumn{1}{|p{2.52cm}}{\centering
38.30-93.58} & 
\multicolumn{1}{|p{2.93cm}|}{\centering
22.6.-70.3} \\ 
\hline
\multicolumn{1}{|p{4.35cm}}{\multirow{3}{*}{\parbox{4.35cm}{\centering
Treatment}}} & 
\multicolumn{1}{|p{2.96cm}}{\centering
Surgery} & 
\multicolumn{1}{|p{2.2cm}}{\centering
56$\%$} & 
\multicolumn{1}{|p{2.52cm}}{\centering
30$\%$} & 
\multicolumn{1}{|p{2.93cm}|}{\centering
40$\%$} \\ 
\hhline{~----}
\multicolumn{1}{|p{4.35cm}}{} & 
\multicolumn{1}{|p{2.96cm}}{\centering
Endovascular} & 
\multicolumn{1}{|p{2.2cm}}{\centering
34$\%$} & 
\multicolumn{1}{|p{2.52cm}}{\centering
35$\%$} & 
\multicolumn{1}{|p{2.93cm}|}{\centering
60$\%$} \\ 
\hhline{~----}
\multicolumn{1}{|p{4.35cm}}{} & 
\multicolumn{1}{|p{2.96cm}}{\centering
None} & 
\multicolumn{1}{|p{2.2cm}}{\centering
10$\%$} & 
\multicolumn{1}{|p{2.52cm}}{\centering
35$\%$} & 
\multicolumn{1}{|p{2.93cm}|}{\centering
0$\%$} \\ 
\hline
\multicolumn{2}{|p{7.31cm}}{Mortality} & 
\multicolumn{1}{|p{2.96cm}}{\centering
30$\%$} & 
\multicolumn{1}{|p{2.2cm}}{\centering
50$\%$} & 
\multicolumn{1}{|p{2.52cm}|}{\centering
10$\%$} \\ 
\hline
\multicolumn{5}{|p{14.959999999999999cm}|}{95$\%$ CI: 95$\%$ confidence interval; HT: hypertension; mFisher: modified Fisher scale; SD: standard deviation; WFNS: World Federation of Neurological Societies scale.} \\ 
\hline
\end{tabular}
\end{adjustbox}
\end{table}
\vspace{1\baselineskip}
The performance of the Swin-UnetR model is presented in Table 2, which shows its robustness across both the validation and testing datasets (Figures 1-3). The model achieved high Dice coefficient, VSI and IoU values, indicating its ability to precisely segment blood regions in NCCT scans (Figure 4). Notably, the model's characteristics extend beyond its segmentation performance. With approximately 62.196 million parameters, the model has the capacity to capture intricate features and patterns within input data. Its computational ability is reflected by its processing speed of approximately 393.340 giga Floating-Point Operations Per Second (GFLOPS), highlighting its suitability for complex tasks. Additionally, the efficiency of the Swin-UnetR model is evident in its inference time of approximately 1.075 seconds per subject, suggesting its potential for real-time or near-real-time applications. These combined attributes underscore the model's ability to provide accurate blood segmentation outcomes while effectively managing computational demands. Principio del formularioFinal del formulario

\begin{table}[H]
\begin{adjustbox}{max width=\textwidth}
\begin{tabular}{p{2.47cm}p{2.31cm}p{3.68cm}p{3.32cm}p{3.19cm}}
\hline
\multicolumn{5}{|p{14.97cm}|}{\textbf{Table 2. The performance of the SWIN-UNETR model for SAH segmentation tasks in the validation, test and external cohorts.}} \\ 
\hline
\multicolumn{1}{|p{2.47cm}}{\centering
\textbf{Sample}} & 
\multicolumn{1}{|p{2.31cm}}{\centering
\textbf{Metric}} & 
\multicolumn{1}{|p{3.68cm}}{\centering
\textbf{Validation}} & 
\multicolumn{1}{|p{3.32cm}}{\centering
\textbf{Test}} & 
\multicolumn{1}{|p{3.19cm}|}{\centering
\textbf{External Test}} \\ 
\hline
\multicolumn{1}{|p{2.47cm}}{\multirow{6}{*}{\parbox{2.47cm}{\centering
Validation}}} & 
\multicolumn{1}{|p{2.31cm}}{\centering
Dice} & 
\multicolumn{1}{|p{3.68cm}}{\centering
0.830 $\pm$ 0.081} & 
\multicolumn{1}{|p{3.32cm}}{\centering
0.873 $\pm$ 0.097} & 
\multicolumn{1}{|p{3.19cm}|}{\centering
0.738 $\pm$ 0.096} \\ 
\hhline{~----}
\multicolumn{1}{|p{2.47cm}}{} & 
\multicolumn{1}{|p{2.31cm}}{\centering
IoU} & 
\multicolumn{1}{|p{3.68cm}}{\centering
0.755 $\pm$ 0.083} & 
\multicolumn{1}{|p{3.32cm}}{\centering
0.810 $\pm$ 0.092} & 
\multicolumn{1}{|p{3.19cm}|}{\centering
0.595 $\pm$ 0.133} \\ 
\hhline{~----}
\multicolumn{1}{|p{2.47cm}}{} & 
\multicolumn{1}{|p{2.31cm}}{\centering
VSI} & 
\multicolumn{1}{|p{3.68cm}}{\centering
0.752 $\pm$ 0.182} & 
\multicolumn{1}{|p{3.32cm}}{\centering
0.840 $\pm$ 0.131} & 
\multicolumn{1}{|p{3.19cm}|}{\centering
0.799 $\pm$ 0.156} \\ 
\hhline{~----}
\multicolumn{1}{|p{2.47cm}}{} & 
\multicolumn{1}{|p{2.31cm}}{\centering
SASD} & 
\multicolumn{1}{|p{3.68cm}}{\centering
2.105 $\pm$ 2.463} & 
\multicolumn{1}{|p{3.32cm}}{\centering
1.866 $\pm$ 2.910} & 
\multicolumn{1}{|p{3.19cm}|}{\centering
1.280 $\pm$ 1.379} \\ 
\hhline{~----}
\multicolumn{1}{|p{2.47cm}}{} & 
\multicolumn{1}{|p{2.31cm}}{\centering
Sensitivity} & 
\multicolumn{1}{|p{3.68cm}}{\centering
0.723 $\pm$ 0.204} & 
\multicolumn{1}{|p{3.32cm}}{\centering
0.821 $\pm$ 0.217} & 
\multicolumn{1}{|p{3.19cm}|}{\centering
0.855 $\pm$ 0.117} \\ 
\hhline{~----}
\multicolumn{1}{|p{2.47cm}}{} & 
\multicolumn{1}{|p{2.31cm}}{\centering
Specificity} & 
\multicolumn{1}{|p{3.68cm}}{\centering
0.997 $\pm$ 0.003} & 
\multicolumn{1}{|p{3.32cm}}{\centering
0.996 $\pm$ 0.004} & 
\multicolumn{1}{|p{3.19cm}|}{\centering
0.997 $\pm$ 0.003} \\ 
\hline
\multicolumn{5}{|p{14.97cm}|}{IoU $=$ intersection over union, VSI $=$ volumetric similarity index, SASD$=$ symmetric average surface distance.} \\ 
\hline
\end{tabular}
\end{adjustbox}
\end{table}
\vspace{1\baselineskip}
\begin{figure}[H]
\includegraphics[width=15.0cm,height=12.76cm]{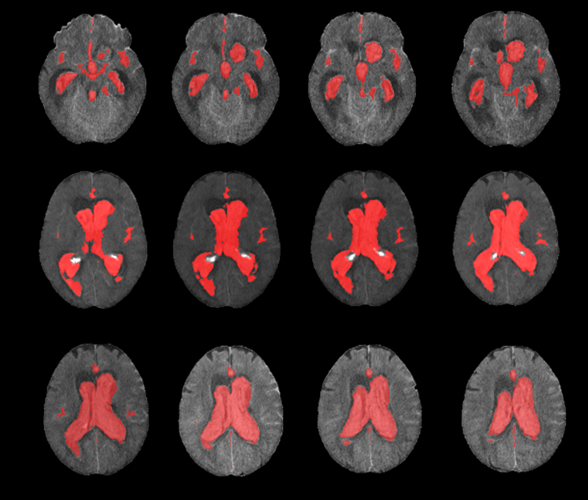}
\end{figure}

\textbf{Figure 1. }NCCT scan of a patient with an aneurysmal SAH grade IV according to the modified Fisher scale. Blood segmentation is represented as a red overlay.

\begin{figure}[H]
\centering
\includegraphics[width=13.02cm,height=21.63cm]{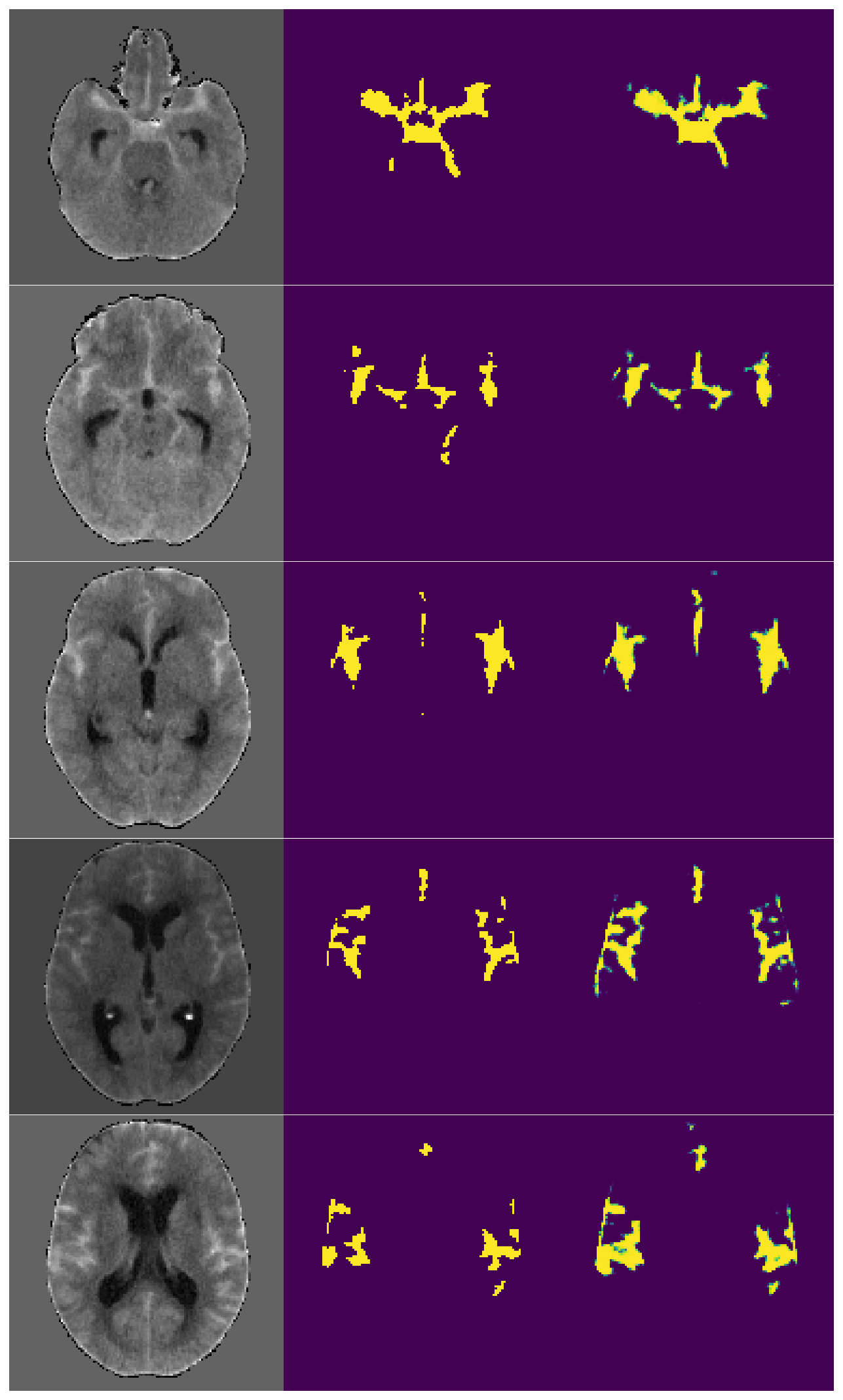}
\end{figure}

\textbf{Figure 2.} NCCT scan slices of a patient with an aneurysmal SAH grade III according to the modified Fisher scale (left column). Manual segmentation of subarachnoid hemorrhage used as ground truth for training and evaluation purposes (central column). Segmentation provided by the Swin-UNETR (right column).

\vspace{1\baselineskip}
\textbf{\begin{figure}[H]
\centering
\includegraphics[width=8.33cm,height=19.72cm]{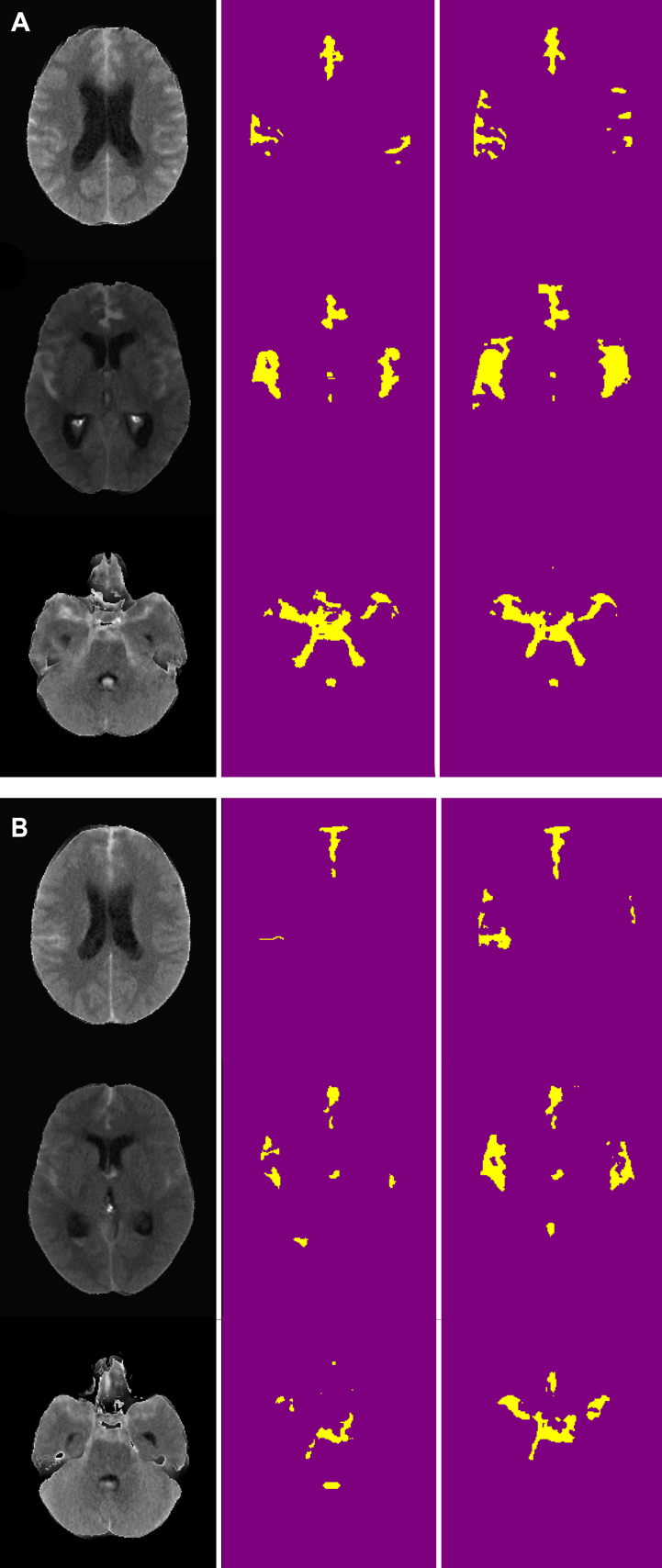}
\end{figure}
}

\textbf{Figure 3.} NCCT scan slices (left column), manual segmentation (central column) and segmentation provided by the model (right column) for two patients with an aneurysmal SAH grade IV according to the modified Fisher scale. A depicts the best concordance between the ground truth and the automatic segmentation, while B represents the lowest degree of concordance in the test cohort.

\vspace{2\baselineskip}
\begin{figure}[H]
\centering
\includegraphics[width=13.5cm,height=13.45cm]{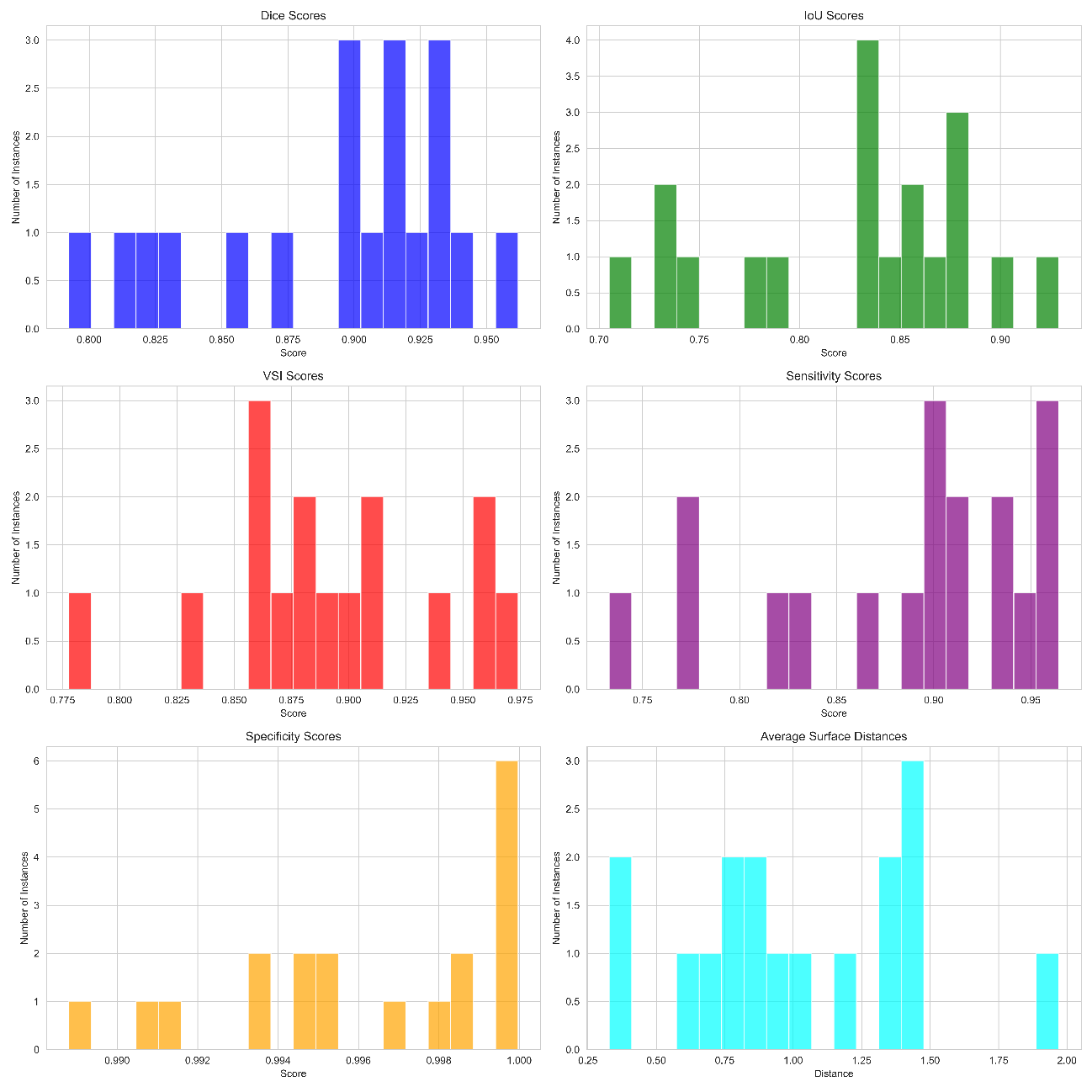}
\end{figure}

\textbf{Figure 4. }Bar chart for every metric used to evaluate the performance of the model. Each bar represents the number of patients sharing the same score for each metric.\textbf{ }IoU: Intersection over Union; VSI: Volumetric similarity index.

\vspace{2\baselineskip}
\textbf{DISCUSSION}

\vspace{1\baselineskip}
In this article, we present a novel model based on the SwinUNETR transformer architecture to accurately perform automatic blood segmentation for patients suffering from spontaneous aneurysmal SAH. The model demonstrated excellent performance across all the implemented evaluation metrics. To enhance the practicality of our model, we created a dedicated repository containing the outcomes of our trained model (\url{https://github.com/smcch/Subarachnoid_Hemorrhage_segmentation_and_mortality_prediction}). This resource is available to encourage individuals to apply this model to their data and further develop its applications and robustness. Our pipeline adheres to a structured approach, encompassing critical image preprocessing stages, such as anonymization, image registration to a normalized spatial context and skull stripping. Consequently, only the original DICOM images from NCCT scans of subjects need to be provided as input. The resulting output includes both the processed NIfTI volume and the NIfTI object of the segmented blood. Furthermore, a PDF was automatically generated with the hemorrhage volume in cm3 in the normalized space, accompanied by CT scan screenshots featuring the segmentation as an overlay (Figure 1). Managing the entire code execution process is effortless through an intuitive graphical interface tailored specifically for the Linux operating system (Figure 5).

\vspace{1\baselineskip}
\begin{figure}[H]
\centering
\includegraphics[width=8.59cm,height=10.37cm]{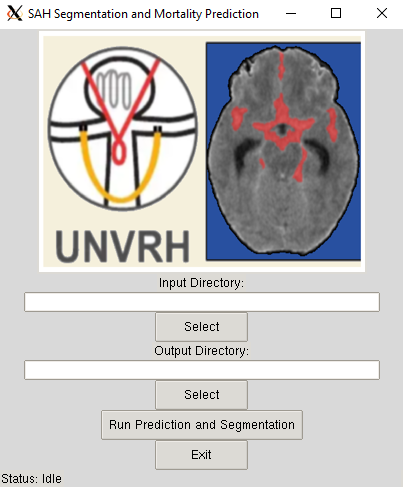}
\end{figure}

\textbf{Figure 5.} The model´s graphical interface in which the input and output directories can be selected initiates the automatic segmentation process.

\vspace{1\baselineskip}
In medical image analysis, UNETR emerged as the first methodology for employing a vision transformer (ViT) as its encoder, diverging from the conventional reliance on a CNN-based feature extractor(19). Previous efforts to incorporate ViT as an additional component within CNN-based architectures outperformed UNETR in terms of accuracy and efficiency across various medical image segmentation tasks (19, 20).

Recently, the concept of the Swin transform was coined to describe a novel approach characterized by its hierarchical architecture, which efficiently computes self-attention through a shifted window partitioning scheme(18). Consequently, Swin Transformers are well suited for downstream tasks because they capitalize on the multiscale features they extract for subsequent processing. Therefore, encouraged by the excellent results of Hamtamizadeh et al. in brain tumor segmentation, we hypothesized that the Swin UNETR could be the optimal tool for blood segmentation in NCCT scans for spontaneous aneurysmal subarachnoid hemorrhage (18).

The development and implementation of an automatic segmentation model for subarachnoid hemorrhage on NCCT images is a challenging task involving multiple hardships. The volumetric nature of NCCT demands a DL algorithm capable of processing a third spatial dimension. Lee et al. suggested that the performance of 3D CNNs might be worse than that of their 2D counterparts(21). This fact might be explained first by the increased number of parameters in 3D CNNs requiring downsampling procedures that lead to the loss of potentially relevant information for the model(22, 23); second by the need for larger datasets to properly train the model as a consequence of dimensionality; and third by the complexity and tediousness of creating a large dataset of skillfully labeled 3D images required as ground truth; finally, all these factors add up to claim greater computational power(24). Transformers overcome some of these challenges. As highlighted by Dosovitskiy et al., despite the overall acceptance and predominance of CNNs in visual recognition tasks, transformers may outperform CNNs while requiring less computational resources(25). This is due to the specific design of transformers, which are designed to capture long-range dependencies in sequences, making them suitable for tasks where contextual information from distant pixels is essential for accurate segmentation. In addition, transformers use self-attention mechanisms to weigh the importance of different parts of an image, focusing on relevant features while processing the entire input; this approach becomes extremely important when segmenting objects with complex shapes such as blood in SAH patients(14, 19). Furthermore, the efficiency and scalability of transformers have allowed them to train models with enormous and unprecedented datasets(26). Other advantages of transformers include their ability to manage multimodal data, improve spatial resolution(27), and parallelize data, indicating their ability to process the input data in parallel due to their self-attention mechanism, thus reducing training and inference times(19).

Recently, Thanellas et al. reported their experience in SAH segmentation using U-NETs as the architecture of a DL algorithm (28). They used a cohort of 96 patients with SAH and a control group of 949 patients whose CT scans were negative for SAH to train and validate their model. Manual segmentation was applied to the CT scans, and individual slices were categorized as either positive or negative for hemorrhage. While some analyses were pixel-based, the primary outcomes of the study were presented on a per-slice and per-patient basis. It has been stated that evaluating a model in terms of pixels is of limited clinical relevance and that slice-based assessment should be the norm (29). However, regarding subarachnoid hemorrhage (SAH), evidence indicates that not only the volume but also the distribution and location of subarachnoid hemorrhage have implications for patient prognosis and the occurrence of neurological complications(5, 9, 11, 30, 31). For this reason, we advocate for the need to create a model capable of providing reliable results at the pixel/voxel level as well. According to our voxel-based analysis, our model outperformed the model developed by Thanellas et al., achieving higher sensitivity (0.82 vs 0.5) and comparable specificity (0.99 vs 1)(28). There might be two factors that increase the accuracy of our model. On the one hand, the transformer architecture has outstanding efficacy in segmentation tasks. On the other hand, as stressed by Lee et al., image preprocessing might have a greater impact on improving the accuracy of the model than the underlying architecture itself(21).

In certain clinical scenarios, precise SAH blood segmentation might be of limited clinical relevance. Certainly, this is not the case for researchers who are leveraging their datasets to extract the utmost clinical information from their NCCT to suggest and validate image-related prognostic indicators. Voxel-based or superior labeling offers two major advantages: first, the required number of annotated images is lower, which allows us to obtain excellent performance even with small datasets(28); second, the accuracy of the model is significantly improved, as demonstrated by Kuo et al. and Chang et al.(32, 33). However, the main drawback of superior labeling is that it is a more demanding and time-consuming task with higher interrater variability than slice-based labeling. Despite this, the actual relevance of rater discrepancies has been deemed limited.

Most articles on this topic use sensitivity and specificity as evaluation metrics; however, our results are conveyed through a range of metrics that comprehensively depict the concordance between the model-generated segmentation and the ground truth (28). Pixel-level agreement was quantified by the DICE score, IOU, and SASD, while the VSI was used to measure volumetric concordance. Therefore, pixel-level agreement is not only gauged as a binary measure but also weighted by the proximity between the ground truth and the predicted segmentation. Indeed, sensitivity and specificity assess the interplay among true positives, false positives, true negatives, and false negatives. In this context, our study provides one of the most thorough characterizations of a segmentation model's performance in detecting intracranial hemorrhage to date(28, 32, 33).

\vspace{1\baselineskip}
\textbf{Limitations}

Our study has several limitations. First, the vast majority of the NCCT scans used to train the model were sourced from a single CT scanner at one institution. Nonetheless, the model demonstrated similar performance on a different subset of patients from another institution. Second, the robustness of the model relies on finely labeled data, in this specific case, on accurately segmented blood on NCCT images. This initial cumbersome and time-consuming task influences the performance of the model. However, our findings corroborate those of other studies suggesting that pixel-based segmentation can offset the need for large datasets (21). Third, the complexity of transformer models may require significant computational resources. The final model can be implemented on a conventional personal computer through a user-friendly interface. Finally, the reliability of the model in clinical practice needs to be tested across new datasets from different institutions and different CT scanners. To facilitate the applicability of the model, we created an open-access repository for researchers and physicians who are willing to apply this model to their own cases and datasets.

\vspace{1\baselineskip}
\textbf{CONCLUSION}

Herein, we present a novel model for automatic blood segmentation in spontaneous SAH patients via NCCT scans based on a transformer approach, specifically the Swin UNETR model. The model, which leverages transformer architectures and self-attention mechanisms, addresses the challenges of segmenting blood in SAH patients and yields high performance across all evaluation metrics. The model shows potential for real-time applications due to its efficiency and computational ability. The present study highlights the importance of precise voxel-level segmentation and discusses the superiority of transformer architectures over traditional CNNs in handling complex segmentation tasks. Finally, the model’s integration into a user-friendly interface enhances its accessibility and potential for widespread clinical application.

\vspace{2\baselineskip}
\textbf{REFERENCES.}

\vspace{1\baselineskip}
\textbf{}1.\ \ \ \ Al-Shahi R, White PM, Davenport RJ, Lindsay KW. Subarachnoid haemorrhage. BMJ. 2006;333(7561):235-40.

2.\ \ \ \ Lo BW, Macdonald RL, Baker A, Levine MA. Clinical outcome prediction in aneurysmal subarachnoid hemorrhage using Bayesian neural networks with fuzzy logic inferences. Comput Math Methods Med. 2013;2013:904860.

3.\ \ \ \ Helbok R, Kurtz P, Vibbert M, Schmidt MJ, Fernandez L, Lantigua H, et al. Early neurological deterioration after subarachnoid hemorrhage: risk factors and impact on outcome. J Neurol Neurosurg Psychiatry. 2013;84(3):266-70.

4.\ \ \ \ Czorlich P, Ricklefs F, Reitz M, Vettorazzi E, Abboud T, Regelsberger J, et al. Impact of intraventricular hemorrhage measured by Graeb and LeRoux score on case fatality risk and chronic hydrocephalus in aneurysmal subarachnoid hemorrhage. Acta Neurochir (Wien). 2015;157(3):409-15.

5.\ \ \ \ Mayfrank L, Hutter BO, Kohorst Y, Kreitschmann-Andermahr I, Rohde V, Thron A, et al. Influence of intraventricular hemorrhage on outcome after rupture of intracranial aneurysm. Neurosurg Rev. 2001;24(4):185-91.

6.\ \ \ \ Zhao B, Yang H, Zheng K, Li Z, Xiong Y, Tan X, et al. Preoperative and postoperative predictors of long-term outcome after endovascular treatment of poor-grade aneurysmal subarachnoid hemorrhage. J Neurosurg. 2017;126(6):1764-71.

7.\ \ \ \ Lagares A, Jimenez-Roldan L, Gomez PA, Munarriz PM, Castano-Leon AM, Cepeda S, et al. Prognostic Value of the Amount of Bleeding After Aneurysmal Subarachnoid Hemorrhage: A Quantitative Volumetric Study. Neurosurgery. 2015;77(6):898-907; discussion

8.\ \ \ \ Teasdale GM, Drake CG, Hunt W, Kassell N, Sano K, Pertuiset B, et al. A universal subarachnoid hemorrhage scale: report of a committee of the World Federation of Neurosurgical Societies. J Neurol Neurosurg Psychiatry. 1988;51(11):1457.

9.\ \ \ \ Sato T, Sasaki T, Sakuma J, Watanabe T, Ichikawa M, Ito E, et al. Quantification of subarachnoid hemorrhage by three-dimensional computed tomography: correlation between hematoma volume and symptomatic vasospasm. Neurol Med Chir (Tokyo). 2011;51(3):187-94.

10.\ \ \ \ Ko SB, Choi HA, Carpenter AM, Helbok R, Schmidt JM, Badjatia N, et al. Quantitative analysis of hemorrhage volume for predicting delayed cerebral ischemia after subarachnoid hemorrhage. Stroke. 2011;42(3):669-74.

11.\ \ \ \ Garcia S, Torne R, Hoyos JA, Rodriguez-Hernandez A, Amaro S, Llull L, et al. Quantitative versus qualitative blood amount assessment as a predictor for shunt-dependent hydrocephalus following aneurysmal subarachnoid hemorrhage. J Neurosurg. 2018;131(6):1743-50.

12.\ \ \ \ Boers AM, Zijlstra IA, Gathier CS, van den Berg R, Slump CH, Marquering HA, et al. Automatic quantification of subarachnoid hemorrhage on noncontrast CT. AJNR Am J Neuroradiol. 2014;35(12):2279-86.

13.\ \ \ \ Jimenez-Roldan L, Alen JF, Gomez PA, Lobato RD, Ramos A, Munarriz PM, et al. Volumetric analysis of subarachnoid hemorrhage: assessment of the reliability of two computerized methods and their comparison with other radiographic scales. J Neurosurg. 2013;118(1):84-93.

14.\ \ \ \ Vaswani A, Shazeer N, Parmar N, Uszkoreit J, Jones L, Gomez AN, et al. Attention is all you need. Advances in neural information processing systems. 2017;30.

15.\ \ \ \ Vandenbroucke JP, von Elm E, Altman DG, Gotzsche PC, Mulrow CD, Pocock SJ, et al. Strengthening the Reporting of Observational Studies in Epidemiology (STROBE): explanation and elaboration. Epidemiology. 2007;18(6):805-35.

16.\ \ \ \ Mongan J, Moy L, Kahn CE, Jr. Checklist for Artificial Intelligence in Medical Imaging (CLAIM): A Guide for Authors and Reviewers. Radiol Artif Intell. 2020;2(2):e200029.

17.\ \ \ \ Yushkevich PA, Yang G, Gerig G. ITK-SNAP: An interactive tool for semiautomatic segmentation of multimodality biomedical images. Annu Int Conf IEEE Eng Med Biol Soc. 2016;2016:3342-5.

18.\ \ \ \ Hatamizadeh A, Nath V, Tang Y, Yang D, Roth HR, Xu D, editors. Swin unetr: Swin transformers for semantic segmentation of brain tumors in mri images. International MICCAI Brainlesion Workshop; 2021: Springer.

19.\ \ \ \ Hatamizadeh A, Tang Y, Nath V, Yang D, Myronenko A, Landman B, et al., editors. Unetr: Transformers for 3d medical image segmentation2022.

20.\ \ \ \ Wang W, Chen C, Ding M, Yu H, Zha S, Li J, editors. Transbts: Multimodal brain tumor segmentation using transformer. Medical Image Computing and Computer Assisted Intervention–MICCAI 2021: 24th International Conference, Strasbourg, France, September 27–October 1, 2021, Proceedings, Part I 24; 2021: Springer.

21.\ \ \ \ Lee H, Yune S, Mansouri M, Kim M, Tajmir SH, Guerrier CE, et al. An explainable deep-learning algorithm for the detection of acute intracranial hemorrhage from small datasets. Nat Biomed Eng. 2019;3(3):173-82.

22.\ \ \ \ Ker J, Singh SP, Bai Y, Rao J, Lim T, Wang L. Image Thresholding Improves 3-Dimensional Convolutional Neural Network Diagnosis of Different Acute Brain Hemorrhages on Computed Tomography Scans. Sensors (Basel). 2019;19(9).

23.\ \ \ \ Yeo M, Tahayori B, Kok HK, Maingard J, Kutaiba N, Russell J, et al. Evaluation of techniques to improve a deep learning algorithm for the automatic detection of intracranial hemorrhage on CT head imaging. Eur Radiol Exp. 2023;7(1):17.

24.\ \ \ \ Arbabshirani MR, Fornwalt BK, Mongelluzzo GJ, Suever JD, Geise BD, Patel AA, et al. Advanced machine learning in action: identification of intracranial hemorrhage on computed tomography scans of the head with clinical workflow integration. NPJ Digit Med. 2018;1:9.

25.\ \ \ \ Dosovitskiy A, Beyer L, Kolesnikov A, Weissenborn D, Zhai X, Unterthiner T, et al. An image is worth 16x16 words: Transformers for image recognition at scale.  .  ICLR2021.

26.\ \ \ \ Lepikhin D, Lee H, Xu Y, Chen D, Firat O, Huang Y, et al. Scaling giant models with conditional computation and automatic sharding. arXiv preprint arXiv:200616668. 2020.

27.\ \ \ \ Raghu M, Unterthiner T, Kornblith S, Zhang C, Dosovitskiy A. Do vision transformers see like convolutional neural networks? Advances in Neural Information Processing Systems. 2021;34:12116-28.

28.\ \ \ \ Thanellas A, Peura H, Lavinto M, Ruokola T, Vieli M, Staartjes VE, et al. Development and External Validation of a Deep Learning Algorithm to Identify and Localize Subarachnoid Hemorrhage on CT Scans. Neurology. 2023;100(12):e1257-e66.

29.\ \ \ \ Thanellas A, Peura H, Wennervirta J, Korja M. Foundations of Brain Image Segmentation: Pearls and Pitfalls in Segmenting Intracranial Blood on Computed Tomography Images. Acta Neurochir Suppl. 2022;134:153-9.

30.\ \ \ \ Schuss P, Hadjiathanasiou A, Brandecker S, Wispel C, Borger V, Guresir A, et al. Risk factors for shunt dependency in patients suffering from spontaneous, nonaneurysmal subarachnoid hemorrhage. Neurosurg Rev. 2019;42(1):139-45.

31.\ \ \ \ Zanaty M, Nakagawa D, Starke RM, Leira EC, Samaniego EA, Guerrero WR, et al. Intraventricular extension of an aneurysmal subarachnoid hemorrhage is an independent predictor of a worse functional outcome. Clin Neurol Neurosurg. 2018;170:67-72.

32.\ \ \ \ Kuo W, Hane C, Mukherjee P, Malik J, Yuh EL. Expert-level detection of acute intracranial hemorrhage on head computed tomography using deep learning. Proc Natl Acad Sci U S A. 2019;116(45):22737-45.

33.\ \ \ \ Chang PD, Kuoy E, Grinband J, Weinberg BD, Thompson M, Homo R, et al. Hybrid 3D/2D Convolutional Neural Network for Hemorrhage Evaluation on Head CT. AJNR Am J Neuroradiol. 2018;39(9):1609-16.

\end{document}